\documentclass{article}

 \usepackage[preprint]{neurips_2026}

\usepackage[utf8]{inputenc} 
\usepackage[T1]{fontenc}    
\usepackage{hyperref}       
\usepackage{url}            
\usepackage{booktabs}       
\usepackage{amsfonts}       
\usepackage{nicefrac}       
\usepackage{microtype}      
\usepackage{xcolor}         

\usepackage{amsmath}
\usepackage[table]{xcolor} 

\usepackage{tcolorbox}
\usepackage{booktabs}
\usepackage{multirow}
\usepackage{subcaption}
\usepackage{float}
\usepackage[capitalize,noabbrev]{cleveref}
\usepackage{wrapfig}

\title{Scaling Pretrained Representations Enables Label-Free Out-of-Distribution Detection Without Fine-Tuning}

\author{%
  Brett Barkley \\
  Department of Electrical and Computer Engineering\\
  The University of Texas at Austin\\
  Austin, TX, USA \\
  \texttt{bbarkley@utexas.edu} \\
  \And
  Preston Culbertson \\
  Department of Computer Science\\
  Cornell University\\
  Ithaca, NY, USA \\
  \texttt{pculbertson@cornell.edu} \\
  \And
  David Fridovich-Keil \\
  Department of Aerospace Engineering and Engineering Mechanics\\
  The University of Texas at Austin\\
  Austin, TX, USA \\
  \texttt{dfk@utexas.edu} \\
}

\begin{document}

\maketitle

\begin{abstract}
Models trained with deep learning often fail to signal when inputs fall outside their training data manifold, leading to unreliable predictions under distribution shift. Prior work suggests that effective out-of-distribution (OOD) detection often requires class-conditional modeling or specialized models obtained through supervised fine-tuning. We revisit this assumption in modern pretrained models and show that their frozen representations already encode sufficient geometric structure for accurate label-free OOD detection. Across 59 backbone–task pairings spanning vision and language, we compare two complementary label-free detectors: a global Mahalanobis estimator fit on unlabeled latent representations, and ReSCOPED, a lightweight, diffusion-based typicality estimator operating on the same features at a local level. Despite their different detection mechanisms, representation scaling reveals a consistent regime-dependent pattern: both local and global detectors' absolute performance improves with better representation quality, and performance gaps between the two detectors disappear across both language and vision tasks as representations scale. These results suggest that label-free OOD detection depends strongly on the geometry exposed by frozen pretrained backbones, reducing the importance of detector choice as backbone scale increases and enabling efficient deployment directly on frozen models. 
\end{abstract}

\begin{figure}[b]
    \centering
    \includegraphics[width=1.0\linewidth]{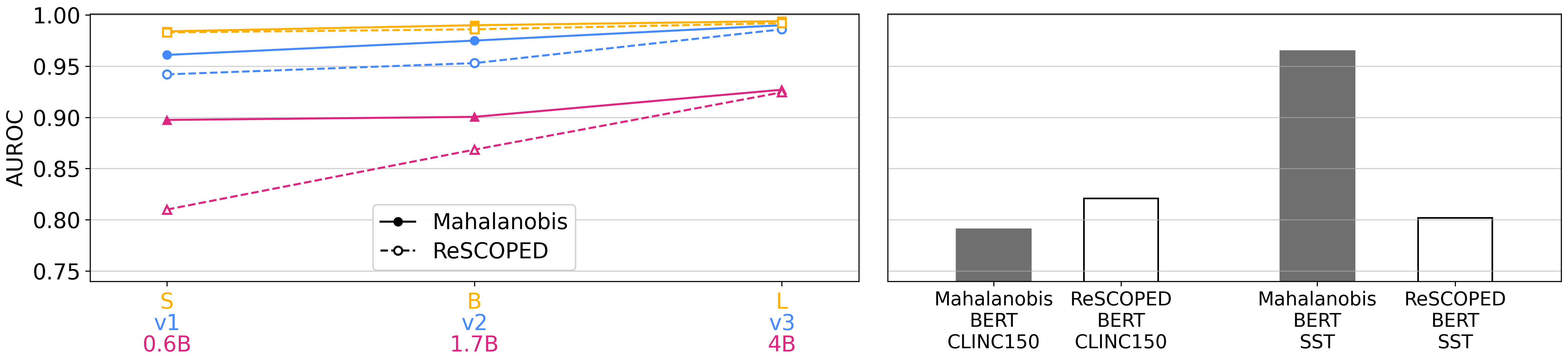}
    \caption{AUROC for Mahalanobis and ReSCOPED across DINO generation, DINOv3 model scale (S, B, L), and Qwen3 model scale (0.6, 1.7, 4.0). For DINO, AUROC is averaged across the 9 tasks in \cref{tab:vision_results_rescoped_priorwork}; for Qwen3, AUROC is averaged across the tasks in \cref{tab:language_results}. Across both vision and language, scaled frozen backbones yield high-performing label-free OOD detectors while also reducing the gap between global Mahalanobis and local score-curvature detectors. In weaker frozen BERT representations, rankings are dataset-dependent: ReSCOPED is stronger on CLINC150, while Mahalanobis is stronger on SST.}

    \label{fig:scaling}
\end{figure}

\section{Introduction}

Whether utilizing a Vision Transformer (ViT) \cite{dosovitskiy2021imageworth16x16words} for image classification or a Large Language Model (LLM) for natural language processing, current architectures generate high confidence predictions even when presented with inputs that fall outside their training manifold. Unsupervised detection strategies are particularly valuable because they rely only on unlabeled in-distribution data and therefore remain task agnostic \citep{liu2020energy, graham2023denoising, heng2024out}. However, traditional out-of-distribution (OOD) methods face persistent challenges: likelihood-based estimators can assign higher probabilities to OOD samples than to training data \citep{nalisnick2018deep, choi2018waic}, while reconstruction-based methods are sensitive to bottleneck tuning \citep{An2015VariationalAB, pinaya2021unsupervised}. In this work, we provide empirical evidence that frozen representation structure governs not only label-free OOD detection performance, but also how much detector choice matters. Here, representation structure refers to how readily OOD-relevant features are exposed to global and local detectors as backbone families and model scales change.

A common perspective across modalities is that strong OOD detection often requires either class labels for in-distribution (ID) data or adaptation of the learned representation. For example, \citet{lee2018simple} model feature representations using class-conditional Gaussian distributions, an assumption that persists in later extensions \citep{liu2024goodllmsoutofdistributiondetection, mueller2025mahalanobisimprovingooddetection, wei2025xmahalanobis}. Similarly, prior work has shown that near-OOD detection remains challenging when using frozen representations, and that supervised fine-tuning can substantially improve performance \citep{fort2021exploring, xu2021unsupervised}. At the same time, OOD detection methods have been shown to exhibit inconsistent performance across datasets, with no single detector consistently outperforming others \citep{tajwar2021truestateoftheartooddetection}. Together, these results suggest that OOD performance depends not only on the detector applied to a representation, but also on the quality and geometry of the representation itself.

This motivates the fully unsupervised setting, where only unlabeled in-distribution data is available. While labels and fine-tuning can improve OOD detection, they limit deployment alongside frozen foundation models: class labels are often unavailable, especially in language settings where models are trained on large-scale unlabeled corpora and deployed on open-ended inputs, and fine-tuning requires maintaining task-adapted model variants. Recent work such as DiffPath has focused on making label-free OOD detection more general-purpose by using a single pretrained diffusion model across multiple tasks \citep{heng2024out}. Recent work has also observed that DINOv2 \citep{oquab2024dinov2learningrobustvisual} and other foundation-model representations can support competitive OOD detection with simple scores such as nearest neighbors \citep{zhang2024openood,zhao2025equippingvisionfoundationmodel}. However, these observations do not establish whether this behavior generalizes across fundamentally different detectors and modalities, nor whether the inconsistencies observed in prior OOD detectors persist as representation quality improves. We therefore ask whether the frozen model already used for the downstream task encodes sufficient representation geometry for label-free OOD detection, avoiding the need for labels or specialized models obtained through supervised fine-tuning.

We study this question using two deliberately different OOD detectors. First, a label-free global Mahalanobis distance-based scheme which models the frozen representation distribution using only unlabeled in-distribution latents, removing the class-conditioning used in prior Mahalanobis-based methods \citep{lee2018simple, liu2024goodllmsoutofdistributiondetection, mueller2025mahalanobisimprovingooddetection, wei2025xmahalanobis}. The second is a typicality-based method, ReSCOPED, that builds upon SCOPED \citep{barkley2025scopedscorecurvatureoutofdistributionproximity} and evaluates local representation geometry using diffusion-based score and curvature information. In particular, diffusion-based detectors can be viewed as multi-scale filters that selectively expose structure at different granularities, allowing them to recover OOD signal when it is not globally accessible. These methods differ substantially in complexity but test the same underlying hypothesis: that modern frozen representations already contain enough OOD-relevant structure for high-performing label-free detection, whether measured through global covariance structure or local score-curvature information.

We validate this representation-centered view across 59 backbone-task pairings sourced from the Bonsai repository \citep{bonsai2024google} and HuggingFace Transformers \citep{wolf2019huggingface}. As shown in \cref{fig:scaling}, stronger representations improve OOD detection as measured by the area under the receiver operating characteristic (AUROC) curve across both vision and language while also reducing the importance of detector choice. In weaker and moderately strong representations, global Mahalanobis and local score-curvature methods can trade places across datasets and backbones. However, as representation quality improves, the performance gap between methods becomes negligible, indicating convergence between global and local views of OOD-relevant geometry.

This regime-dependent pattern appears in both vision and language: detector choice matters more in weaker or less structured representations, while in strong modern backbones the performance gap between simple global and expressive local methods is small and both achieve strong performance. Because both detectors operate directly on latents from frozen models already used downstream, this representation-centered view also enables efficient deployment. For example, ReSCOPED can act as a lightweight LLM gatekeeper prior to decoding, adding only 5.4 ms of latency for Qwen3-4B \citep{yang2025qwen3technicalreport}. Our ablations further show that the highest performing score-curvature detection occurs across a broad and consistent diffusion-noise plateau, suggesting that OOD-relevant structure is exposed at stable scales across Transformer \citep{vaswani2017attention} backbones.

\begin{tcolorbox}[leftrule=1.5mm,top=1mm,bottom=0mm]
\textbf{Key Contributions:}
\begin{enumerate}
\item We show that modern pretrained vision and language representations enable strong label-free OOD detection without fine-tuning or class labels, using ReSCOPED and Mahalanobis detectors to compare local/global views of representation geometry.
\item We show that backbone family and scale shape detector choice: in earlier or smaller backbones, no method is universally best, while DINOv3 and Qwen3 scaling make the gap between global and local OOD detectors negligible.
\item We identify a consistent denoising-scale plateau across vision and language Transformers, suggesting OOD-relevant structure becomes accessible at a shared scale.
\item We demonstrate a practical deployment pathway via low-latency OOD detection on frozen models, including real-time LLM prefill-stage gatekeeping.
\end{enumerate}
\end{tcolorbox}

\section{Background and Related Work}
\subsection{Representation Geometry and OOD Detection}

Out-of-distribution (OOD) detection can be viewed as probing the geometry of a model's learned representation. Let $z = \phi(x) \in \mathbb{R}^d$ denote the representation of an input $x$ under a frozen encoder $\phi$. Different OOD methods vary in what information they require and what geometry they measure: distance-based methods estimate global or class-conditional structure \citep{lee2018simple, mueller2025mahalanobisimprovingooddetection, wei2025xmahalanobis}, score-based typicality methods \citep{heng2024out,barkley2025scopedscorecurvatureoutofdistributionproximity} capture local quantities such as gradients and curvature, and supervised or fine-tuned approaches \citep{xu2021unsupervised, fort2021exploring} modify $\phi$ itself to expose OOD-relevant structure.

The class-conditional formulation of \citet{lee2018simple} considers a labeled in-distribution dataset $\{(x_i, y_i)\}_{i=1}^N$, where $x_i$ is an input and $y_i \in \{1, \dots, C\}$ is its class label. Let $N_c$ denote the number of samples in class $c$, and let $\mathcal{I}_c = \{i : y_i = c\}$ denote the corresponding class index set. The method estimates per-class means $\mu_c = \frac{1}{N_c} \sum_{i \in \mathcal{I}_c} z_i$ and a shared covariance $\Sigma_{\mathrm{class}} = \frac{1}{N} \sum_{c=1}^C \sum_{i \in \mathcal{I}_c} (z_i - \mu_c)(z_i - \mu_c)^\top$. The OOD score is then defined as the minimum Mahalanobis distance to the class means: $S_{\mathrm{class}}(z) = \min_c (z - \mu_c)^\top \Sigma_{\mathrm{class}}^{-1} (z - \mu_c)$.

Later extensions continue to leverage class-conditional structure in feature space \citep{mueller2025mahalanobisimprovingooddetection, wei2025xmahalanobis}. Prior work has also shown that OOD detector rankings can be inconsistent across datasets, with no single method uniformly outperforming others \citep{tajwar2021truestateoftheartooddetection}. Recent work has further shown that the performance of OOD detectors such as Mahalanobis distance is strongly influenced by representation geometry, including spectral properties and intrinsic dimensionality \citep{janiak2026geometrybasedviewmahalanobisood}. Our work connects these observations by showing that representation geometry shapes detector choice: when OOD-relevant structure is weak or only partially organized, the best-performing method can depend on the dataset, while scaled pretrained backbones can make the same structure accessible to both local and global methods.

In contrast to class-conditional methods, we focus on the fully label-free setting, where detectors are fit using only unlabeled in-distribution data and do not have access to additional class labels. This setting often arises when deploying frozen models on unlabeled task data or user prompts. In this setting, we compare two complementary probes of the same frozen representation: a global covariance probe and a local score-curvature probe.

\subsection{Score-Based Typicality and Local Geometry}

Models can assign higher likelihood to OOD samples than to in-distribution (ID) data \citep{nalisnick2018deep, choi2018waic} because probability mass concentrates not at the mode, but in a thin shell known as the typical set \citep{cover2006elements, nalisnick2019detecting}. Consequently, high-likelihood samples may still be atypical under the true data distribution, motivating alternatives to direct density thresholding \citep{ren2019likelihood}, including supervised fine-tuning or outlier-exposure methods \citep{hendrycks2016baseline, lee2018simple, wei2025xmahalanobis} and unsupervised latent-space methods based on frozen representations \citep{xu2021unsupervised, gulati2024out}.

Score-based diffusion models \citep{sohl2015deep, song2020score} learn a time-dependent score function $s_\theta(x_t, t) \approx \nabla_x \log p_t(x_t)$ under progressive noise perturbations. The model is trained to reverse a Gaussian noising process, and $p_t$ is the distribution of data at stage $t$ of the diffusion process; the learned score field can be used both for generation and to estimate local geometric quantities such as curvature.

One way to incorporate local geometry is through the score-curvature ratio introduced by \citet{barkley2025scopedscorecurvatureoutofdistributionproximity}, motivated by the Fisher information identity. Let $p(x)$ denote the in-distribution density and $s(x) = \nabla_x \log p(x)$ its score function. Then $\mathbb{E}_{x \sim p}[\|s(x)\|^2] = \mathbb{E}_{x \sim p}[-\operatorname{Tr}(\nabla_x s(x))]$, which motivates the typicality statistic $T(x) = \|s(x)\|^2 / \kappa(x)$, where $\kappa(x) = -\operatorname{Tr}(\nabla_x s(x))$. For $d$-dimensional standard Gaussian $p$, typical samples $x \sim p$ concentrate near $\|x\| = \sqrt{d}$ and satisfy $T(x) \approx 1$. For non-Gaussian data, $T(x)$ need not equal 1 for all points of the typical set; however, empirically, diffusion score fields associated with non-Gaussian data produce score-curvature statistics that have a consistent value for in-distribution samples and deviate for OOD samples, providing an unsupervised typicality signal.

\section{Global and Local Probes of Representation Geometry}
\label{sec:probes}

We compare two label-free probes of the same frozen representation distribution: global Mahalanobis distance fit to unlabeled in-distribution representations, and ReSCOPED, a local score-curvature detector introduced in this work. Together, they test whether OOD-relevant structure is accessible through both global covariance geometry and local typicality estimates.

\subsection{Global Mahalanobis Probe}

Assume access to unlabeled in-distribution inputs $\{x_i\}_{i=1}^N$ and a frozen encoder $\phi$, let $z_i=\phi(x_i)$. Without labels, class means cannot be estimated. We therefore fit a global Gaussian model to the representation distribution, with mean $\mu = \frac{1}{N}\sum_{i=1}^N z_i$ and covariance $\Sigma_{\mathrm{global}} = \frac{1}{N}\sum_{i=1}^N (z_i-\mu)(z_i-\mu)^\top + \lambda I$, where $\lambda>0$ is a regularization parameter. The OOD score is $S_{\mathrm{global}}(z) = (z-\mu)^\top \Sigma_{\mathrm{global}}^{-1}(z-\mu)$. This provides a label-free global probe of representation geometry.

\subsection{Local Score-Curvature Probe: ReSCOPED}

We use ReSCOPED, a representation-based variant of SCOPED \cite{barkley2025scopedscorecurvatureoutofdistributionproximity}, as the local counterpart to the global Mahalanobis method. Rather than modeling the full representation distribution with a single covariance, ReSCOPED trains a lightweight diffusion observer on frozen representations and evaluates local score-curvature typicality.

Formally, given an input $x$, we extract a representation $z = \phi(x)$ from a frozen encoder $\phi$, and train an Elucidated Diffusion Model (EDM) $s_\theta(z,\sigma)$ \citep{karras2022elucidating} on these representations. At test time, we compute the SCOPED typicality statistic \citep{barkley2025scopedscorecurvatureoutofdistributionproximity}:
\[
T(z) = \frac{\|s_\theta(z, \sigma)\|^2}{\kappa(z)}, \quad \kappa(z) = -\operatorname{Tr}(\nabla_z s_\theta(z, \sigma)),
\]
where the trace is approximated using Hutchinson’s estimator \citep{hutchinson1989stochastic}.

We obtain anomaly scores by fitting a Kernel Density Estimator (KDE) to in-distribution values of $T(z)$ and evaluating the negative log-likelihood at test time. A noise level ($\sigma \approx 0.09$) is selected via a denoising sweep on a validation task (SVHN vs.\ CelebA) and used across all experiments, avoiding per-dataset tuning required by prior diffusion-based approaches \citep{heng2024out, barkley2025scopedscorecurvatureoutofdistributionproximity, ding2025revisiting}. This value lies within a broad plateau of highest performing noise levels for OOD detection observed across models and modalities, suggesting that OOD-relevant structure in pretrained Transformer \citep{vaswani2017attention} representations becomes accessible within a stable denoising range. We analyze this phenomenon further in \Cref{sec:transformergeometry}.

By combining a frozen pretrained encoder (e.g., ViT, DINOv3, or Qwen3) with a lightweight diffusion-based observer (approximately 8M parameters), ReSCOPED avoids full-model fine-tuning, which is commonly required in prior OOD detection approaches. Training completes in two minutes on a single GPU, and inference requires one encoder forward pass, one observer forward pass, and a single Jacobian-vector product, corresponding to approximately 5.4 ms of additional latency for a 2560-dimensional latent from Qwen3-4B \citep{yang2025qwen3technicalreport} on an NVIDIA 4090.

\begin{wrapfigure}{r}{0.45\textwidth}
    \vspace{-10pt}
    \centering
    \includegraphics[width=\linewidth]{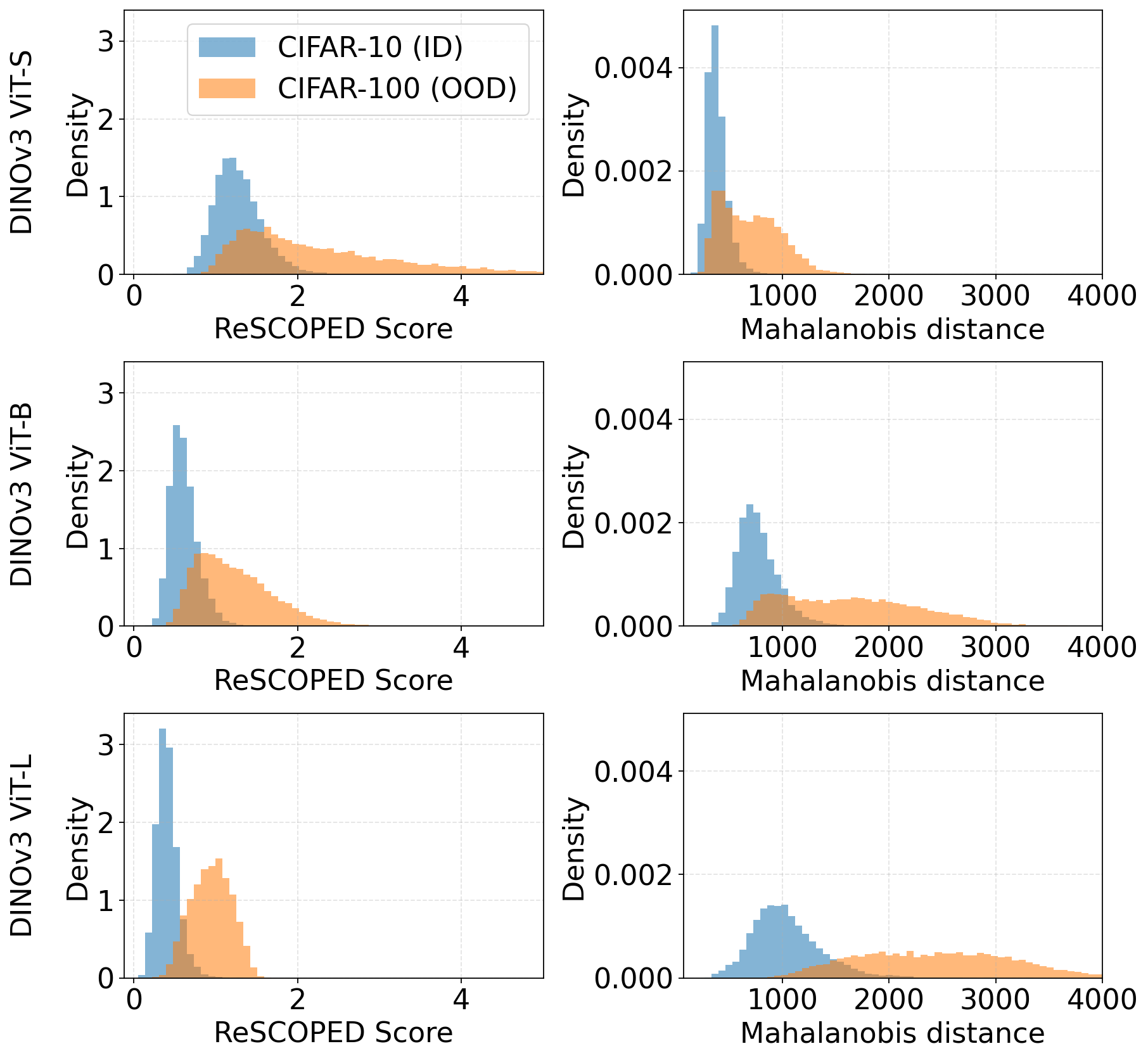}
    \vspace{-8pt}
    \caption{ID/OOD score distributions for CIFAR-10 (ID) vs.\ CIFAR-100 (OOD) across DINOv3 scale (rows: ViT-S/B/L; columns: ReSCOPED and Mahalanobis scores). Both methods improve with model scale. }
\label{fig:local_global_separability}
    \vspace{-10pt}
\end{wrapfigure}

Together, these two methods provide complementary views of the same representation distribution: Mahalanobis measures global covariance structure, while ReSCOPED measures local score-curvature typicality. \Cref{fig:local_global_separability} provides a qualitative preview of this contrast. As the backbone improves, ReSCOPED scores become sharper with reduced ID/OOD overlap, while Mahalanobis separation improves because OOD samples move farther outward in global radius even though the ID distribution broadens. Thus, stronger representations do not simply make the data more Gaussian; they make OOD-relevant deviations more visible to both local and global methods.

\section{Vision and Language Processing Results}

\paragraph{Vision benchmarks.}
We evaluate OOD detection on CIFAR-10 (C10) \citep{Krizhevsky2009LearningML}, SVHN \citep{Netzer2011ReadingDI}, CelebA \citep{liu2015faceattributes}, and CIFAR-100 (C100). For harder near-OOD tasks, we additionally evaluate ImageNet-200 as ID against SSB-hard and NINCO as OOD, following the OpenOOD v1.5 benchmark protocol \citep{zhang2024openood}. Since the small-scale datasets have differing resolutions, we follow \citet{heng2024out} and apply a resizing procedure that avoids artifacts that could trivially simplify OOD detection. Specifically, all images are first downsampled to the lowest native resolution among the datasets (32$\times$32 for CIFAR and SVHN), and then upsampled to the model input resolution.

We compare against generative baselines including SCOPED \citep{barkley2025scopedscorecurvatureoutofdistributionproximity}, DiffPath \citep{heng2024out}, and Improved CD \citep{du2021improved}, as well as likelihood-based methods including Density of States Estimation (DoSE) \citep{morningstar2021density}, the typicality test of \citet{nalisnick2019detecting}, WAIC \citep{choi2018waic}, and the likelihood-ratio score of \citet{ren2019likelihood}. We also include diffusion-based baselines such as NLL (negative ELBO), DDPM-OOD \citep{graham2023denoising}, LMD \citep{liu2023unsupervised}, and MSMA \citep{mahmood2020multiscale}. Following prior work \citep{barkley2025scopedscorecurvatureoutofdistributionproximity}, we report AUROC and computational cost in terms of the number of forward passes and Jacobian-vector products (JVPs), which together correspond to the total number of function evaluations for diffusion-based methods.

\paragraph{Language benchmarks.}
For language tasks, we evaluate on CLINC150 \citep{larson2019evaluation} and a cross-corpus SST benchmark following \citet{xu2021unsupervised}. CLINC150 evaluates cross-intent OOD detection, while SST evaluates cross-corpus OOD detection using out-of-domain corpora. Following \citet{xu2021unsupervised}, we report AUROC, detection accuracy (DTACC), and area under the precision-recall curve for both in-domain (AUIN) and out-of-domain (AUOUT) samples. All baselines use a frozen BERT \citep{devlin2019bert} backbone and differ in how in-distribution representations are used for OOD detection. This includes methods from \citet{xu2021unsupervised} which covers frozen feature-based methods like (class-conditional) Euclidean-based and Mahalanobis-based distance functions (EDF and MDF, respectively) and approaches that adapt representations using in-distribution data without access to task-specific OOD or class labels like Binary Classification with Auxiliary Dataset (BCAD) and In-Domain Masked Language Modeling (IMLM). We also evaluate EC-NNK-Means \citep{gulati2024out} under the frozen BERT setting to ensure consistency with prior label-blind evaluations. 

\subsection{Near-OOD Detection Without Labels or Fine-Tuning}
\label{sec:finetuneood}

We first isolate a concrete question raised by prior near-OOD work: whether strong near-OOD detection requires supervised adaptation of the representation. \citet{fort2021exploring} argue that full-parameter supervised fine-tuning of large Vision Transformers is important for strong near-OOD performance. Their evaluation uses class-conditional Mahalanobis detectors, which rely on labels to define the OOD score \citep{lee2018simple, mueller2025mahalanobisimprovingooddetection, wei2025xmahalanobis}. We test whether modern frozen representations can close this gap in the fully label-free setting, where the encoder is fixed and only unlabeled in-distribution representations are available.

\begin{wrapfigure}{r}{0.65\textwidth}
    \vspace{-10pt}
    \centering
    \begin{subfigure}[b]{0.32\textwidth}
        \centering
        \includegraphics[width=\linewidth, height=0.18\textheight, keepaspectratio]{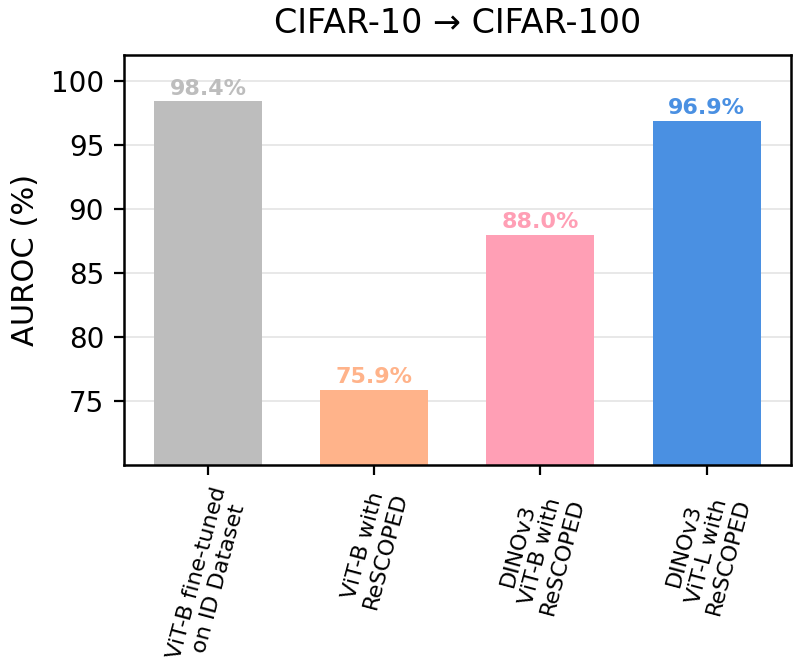}
        \vspace{-8pt}
        \caption{CIFAR-10 to CIFAR-100}
        \label{fig:cifar10_to_100}
    \end{subfigure}
    \hfill
    \begin{subfigure}[b]{0.32\textwidth}
        \centering
        \includegraphics[width=\linewidth, height=0.18\textheight, keepaspectratio]{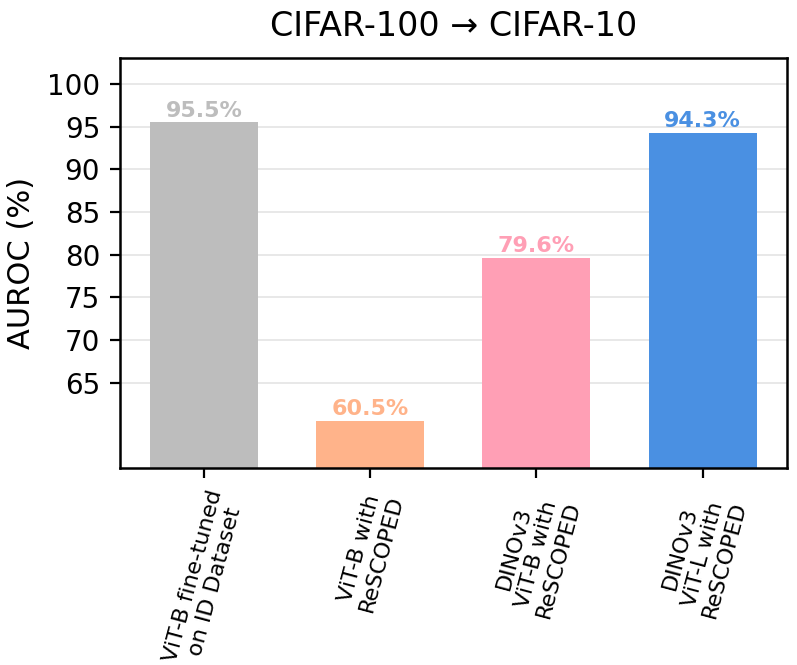}
        \vspace{-8pt}
        \caption{CIFAR-100 to CIFAR-10}
        \label{fig:cifar100_to_10}
    \end{subfigure}
    \caption{Comparison of ReSCOPED AUROC performance using frozen encoders of varying representational quality against fine-tuning methods across near-OOD datasets.}
    \label{fig:finetune_auroc}
    \vspace{-10pt}
\end{wrapfigure}

As shown in \cref{fig:finetune_auroc}, representations from modern frozen backbones achieve competitive near-OOD performance when paired with ReSCOPED. For example, on CIFAR-100 (ID) versus CIFAR-10 (near-OOD), ReSCOPED achieves 94.3\% AUROC using a frozen DINOv3-L backbone \citep{simeoni2025dinov3}, closely matching fine-tuned models \citep{fort2021exploring}. This suggests that modern frozen representations can already expose OOD-relevant geometry, even in near-OOD settings where the ID and OOD distributions are similar. We next ask whether this structure is specific to ReSCOPED on vision tasks, or whether it also appears for simpler global detectors and in other modalities.

\subsection{Main Vision Results}

Our vision experiments ask whether the complexity of recent OOD detectors is justified when operating on modern frozen representations. Here, we study three empirical questions. First, how well can simple methods perform for OOD detection with modern representations, compared to methods using class labels or fine-tuning? Second, at smaller model scales, is relative detector performance dataset-dependent? Third, as modern pretrained representations scale, does the gap between simple global detectors (Mahalanobis) and more sophisticated local detectors (ReSCOPED) shrink? If so, this would suggest that representation geometry, rather than detector complexity alone, drives OOD performance in modern frozen backbones.

The first question is answered in \cref{tab:vision_results_rescoped}: modern frozen DINO representations enable high-performing label-free OOD detection across the small-scale vision benchmarks. ReSCOPED achieves strong AUROC while requiring far fewer function evaluations than prior diffusion-based methods. However, the DINOv3 results also answer the third question: this performance is not unique to ReSCOPED, since a simple label-free Mahalanobis detector, fit only on unlabeled representations, matches or slightly exceeds ReSCOPED across DINOv3 scales (S, B, L). This suggests that in this regime, performance is governed less by detector complexity and more by the geometry of the frozen representation, which makes OOD-relevant structure accessible to both global and local detectors.

\begin{table}[]
\begin{center}
\caption{Performance comparison across vision in- vs out-of-distribution detection tasks as measured by AUROC score. Higher is better. \textbf{Bold} is best and \underline{underline} is second best. The first set of baselines and the reporting format are reproduced from \citet{barkley2025scopedscorecurvatureoutofdistributionproximity}. We additionally report OpenOOD's ImageNet-200 ID results against SSB-hard and NINCO OOD datasets for the DINOv3 methods. As in prior work, we report computational cost as \#\texttt{F} + \#\texttt{J}, where \texttt{F} denotes a forward pass and \texttt{J} a Jacobian--vector product.}
\label{tab:vision_results_rescoped}
\resizebox{1.0\textwidth}{!}{%
\begin{tabular}{lccccccc|ccc c}
\toprule
\textbf{Method} 
& \multicolumn{3}{c}{\textbf{C10 vs}} 
& \multicolumn{3}{c}{\textbf{CelebA vs}} 
& \textbf{Avg} 
& \multicolumn{3}{c}{\textbf{ImageNet-200 vs}} 
& \textbf{\#\texttt{F} + \#\texttt{J}} \\
 & SVHN & CelebA & C100 
 & C10 & SVHN & C100 
 & 
 & SSB-hard & NINCO & Avg 
 & \\
\cmidrule(lr){2-4} \cmidrule(lr){5-7} \cmidrule(lr){9-11}
Improved CD \citep{du2021improved} & 0.910 & -- & 0.830 & -- & -- & -- & -- & -- & -- & -- & -- \\
DoSE \citep{morningstar2021density} & 0.955 & 0.995 & 0.571 & 0.949 & 0.997 & 0.956 & 0.904 & -- & -- & -- & -- \\
WAIC \citep{choi2018waic} & 0.143 & 0.928 & 0.532 & 0.507 & 0.139 & 0.535 & 0.464 & -- & -- & -- & -- \\
TT \citep{nalisnick2019detecting} & 0.870 & 0.848 & 0.548 & 0.634 & 0.982 & 0.671 & 0.759 & -- & -- & -- & -- \\
LR \citep{ren2019likelihood} & 0.064 & 0.914 & 0.520 & 0.323 & 0.028 & 0.357 & 0.368 & -- & -- & -- & -- \\
\midrule
\multicolumn{12}{c}{\textbf{Diffusion-based}} \\
\midrule
NLL & 0.091 & 0.574 & 0.521 & 0.814 & 0.105 & 0.786 & 0.482 & -- & -- & -- & 1000\texttt{F} + 0\texttt{J} \\
MSMA \citep{mahmood2020multiscale} & 0.957 & \textbf{1.000} & 0.615 & 0.910 & 0.996 & 0.927 & 0.901 & -- & -- & -- & 10\texttt{F} + 0\texttt{J} \\
DDPM-OOD \citep{graham2023denoising} & 0.390 & 0.659 & 0.536 & 0.795 & 0.636 & 0.778 & 0.632 & -- & -- & -- & 350\texttt{F} + 0\texttt{J} \\
LMD \citep{liu2023unsupervised} & 0.992 & 0.557 & 0.604 & 0.989 & \textbf{1.000} & 0.979 & 0.854 & -- & -- & -- & $10^4$\texttt{F} + 0\texttt{J} \\
\midrule
\multicolumn{12}{c}{\textbf{Curvature and Diffusion-Based}} \\
\midrule
DiffPath \citep{heng2024out} & 0.910 & 0.897 & 0.590 & 0.998 & \textbf{1.000} & \textbf{0.998} & 0.899 & -- & -- & -- & 10\texttt{F} + 0\texttt{J} \\
SCOPED \citep{barkley2025scopedscorecurvatureoutofdistributionproximity} & 0.814 & 0.940 & 0.477 & 0.925 & 0.994 & 0.962 & 0.852 & -- & -- & -- & 2\texttt{F} + 2\texttt{J} \\
\midrule
\midrule
\emph{Label-Free Mahalanobis DINOv3 ViT-S} & \textbf{0.997} & \textbf{1.000} & 0.864 & \textbf{1.000} & \textbf{1.000} & \textbf{0.998} & 0.977 & 0.714 & 0.700 & 0.707 & \textbf{1\texttt{F} + 0\texttt{J}} \\
\emph{Label-Free Mahalanobis DINOv3 ViT-B} & 0.990 & \textbf{1.000} & 0.923 & \textbf{1.000} & \textbf{1.000} & \underline{0.997} & 0.985 & 0.795 & 0.800 & 0.797 & \textbf{1\texttt{F} + 0\texttt{J}} \\
\emph{Label-Free Mahalanobis DINOv3 ViT-L} & 0.983 & \textbf{1.000} & \textbf{0.970} & \textbf{1.000} & \underline{0.999} & \underline{0.997} & \textbf{0.991} & \textbf{0.865} & 0.908 & \underline{0.887} & \textbf{1\texttt{F} + 0\texttt{J}} \\
\emph{ReSCOPED DINOv3 ViT-S} & \underline{0.994} & \underline{0.999} & 0.864 & \underline{0.999} & \textbf{1.000} & 0.993 & 0.975 & 0.617 & 0.612 & 0.614 & \underline{2\texttt{F} + 1\texttt{J}} \\
\emph{ReSCOPED DINOv3 ViT-B} & 0.968 & 0.996 & 0.919 & \textbf{1.000} & \textbf{1.000} & 0.996 & 0.980 & 0.829 & 0.838 & 0.834 & \underline{2\texttt{F} + 1\texttt{J}} \\
\emph{ReSCOPED DINOv3 ViT-L} & 0.969 & 0.997 & \underline{0.969} & \textbf{1.000} & 0.997 & 0.995 & \underline{0.988} & \underline{0.855} & \textbf{0.910} & 0.882 & \underline{2\texttt{F} + 1\texttt{J}} \\
\bottomrule
\end{tabular}%
}
\end{center}
\vspace{-22pt}
\end{table}

To test whether this convergence is merely an artifact of saturated benchmarks, \cref{tab:vision_results_rescoped} also includes harder OpenOOD's \citep{zhang2024openood} ImageNet-200 near-OOD evaluations against SSB-hard and NINCO. These settings are substantially less saturated: DINOv3-L label-free Mahalanobis obtains 0.865 AUROC on SSB-hard and 0.908 on NINCO, compared with near-perfect AUROC on many smaller reproduced benchmarks. Nevertheless, the same scale-dependent pattern persists. On ImageNet-200, the average AUROC gap between Mahalanobis and ReSCOPED decreases from 0.093 for DINOv3-S and 0.037 for DINOv3-B to only 0.005 for DINOv3-L. Thus, the harder OpenOOD results support detector convergence even when absolute detection remains nontrivial.

Finally, these results address the second question by connecting detector variability to scale. Prior work shows that detector rankings can depend strongly on the structure of the ID and OOD datasets, with no method consistently dominating across dataset pairs \citep{tajwar2021truestateoftheartooddetection}. For small models, our results match this prediction of noisy rankings. At larger scales, however, the performance converges, suggesting these models learn representations that even simple detectors can exploit. This convergence foreshadows the empirical regularities discussed in \cref{sec:transformergeometry}, where we show that Transformer representations expose OOD-relevant structure across a stable range of denoising scales.

\subsection{Language Results}
\label{sec:main_language}

\begin{table}[b]
 \setlength{\tabcolsep}{3pt}
		\centering
\scalebox{.7}{
\begin{tabular}{l|c|cccc|cccc}
\toprule
 &  & \multicolumn{4}{c|}{CLINC150} & \multicolumn{4}{c}{SST} \\ 
& \small\# features 
& \footnotesize AUROC &  \footnotesize DTACC & \footnotesize AUIN & \footnotesize AUOUT 
& \footnotesize AUROC &  \footnotesize DTACC & \footnotesize AUIN & \footnotesize AUOUT \\ 
\midrule\midrule 
\multicolumn{10}{c}{\small \textit{Frozen Backbone Approaches}} \\ \midrule
EDF  & 12	&  0.553 & 0.552  & 0.843 &  0.203 & 0.901 & 0.848 & 0.928 & 0.842 \\
MDF            & 12      & 0.767 & 0.711 &  0.934 & 0.382 & 0.933 & 0.875 & 0.949 & 0.891 \\
EC-NNK-Means                 & 768      & 0.672 & 0.818 & 0.895 & 0.289 & -- & -- & -- & -- \\
\midrule
\emph{BERT + Label-free Mahalanobis + SMP} & 1 & 0.791 & 0.828 & 0.939 & 0.447 & 0.965 & 0.907 & 0.972 & 0.953 \\
\emph{BERT + ReSCOPED + SMP} & 1 & 0.815 & 0.828 & 0.949 & 0.463 & 0.802 & 0.730 & 0.822 & 0.778 \\

\midrule
\multicolumn{10}{c}{\small \textit{Fine-Tuning Approaches}} \\ \midrule
IMLM +  MDF      & 12       & 0.778 & 0.722 & 0.938 & 0.391 & 0.936 & 0.881 & 0.975 & 0.894 \\
BCAD +  MDF    & 12    & 0.812  &	0.745  &	0.946  &	0.474 & \underline{0.970} & \underline{0.945} & \underline{0.980} & \underline{0.948} \\
IMLM + BCAD +  MDF & 12 & 0.821  &	0.756  &	0.950  &	0.476 & \textbf{0.981} & \textbf{0.954} & \textbf{0.987} & \textbf{0.959} \\
\midrule
\rowcolor[HTML]{FFEBCC} \multicolumn{10}{c}{\small \textit{Scaling Behavior (Modern Backbones)}} \\ \midrule
\emph{Qwen3-0.6B + Label-free Mahalanobis + SMP} & 1 & 0.856 & \underline{0.859} & 0.960 & 0.617 & 0.939 & 0.871 & 0.941 & 0.938 \\
\emph{Qwen3-1.7B + Label-free Mahalanobis + SMP} & 1 & 0.867 & 0.858 & 0.965 & 0.614 & 0.934 & 0.859 & 0.941 & 0.928 \\
\emph{Qwen3-4.0B + Label-free Mahalanobis + SMP} & 1 & \textbf{0.903} & \textbf{0.872} & \textbf{0.976} & \textbf{0.683} & 0.951 & 0.885 & 0.954 & 0.947 \\
\emph{Qwen3-0.6B + ReSCOPED + SMP} & 1 & 0.856 & 0.825 & 0.951 & 0.643 & 0.764 & 0.698 & 0.757 & 0.763 \\
\emph{Qwen3-1.7B + ReSCOPED + SMP} & 1 & 0.861 & 0.856 & 0.963 & 0.599 & 0.876 & 0.811 & 0.886 & 0.853 \\
\emph{Qwen3-4.0B + ReSCOPED + SMP} & 1 & \underline{0.900} & \textbf{0.872} & \underline{0.974} & \underline{0.669} & 0.949 & 0.881 & 0.952 & 0.945 \\
\bottomrule
\end{tabular}
}
\caption{OOD detection performance on CLINC150 and SST. Larger values indicate better performance. BERT baselines are reproduced from \citet{xu2021unsupervised}. The Qwen3 rows evaluate scaling behavior in modern frozen language backbones using label-free Mahalanobis and ReSCOPED on mean-pooled hidden representations.}

\label{tab:language_results}
\vspace{-6pt}

\end{table}

Language provides a cross-modal test of our representation-centered hypothesis: if the regime-dependent behavior observed in our vision results reflects representation geometry rather than vision-specific structure, then weaker frozen language representations should also show dataset-dependent detector rankings, while modern scaled language backbones should reduce the gap between detector types. For both Mahalanobis and ReSCOPED detectors, we use a standard sequence mean pooling (SMP) strategy, averaging token-level hidden states into a single utterance representation before applying each detector. As summarized in \cref{tab:language_results}, the BERT results support the first prediction. On CLINC150, ReSCOPED improves over the label-free Mahalanobis detector using the same SMP representation, while on the SST benchmark, label-free Mahalanobis performs better. Thus, in weaker frozen language representations, no single detector is consistently better across datasets.

The Qwen3 results provide a cross-modal confirmation of the scaling behavior observed in vision and show that scaled language backbones reduce the performance gap between global covariance and local score-curvature detectors. On CLINC150, label-free Mahalanobis and ReSCOPED are closely matched across Qwen3 scales. On SST, ReSCOPED initially lags behind Mahalanobis at smaller Qwen3 scales, but closely matches it once Qwen3-4B is used. Thus, scaled language representations make OOD-relevant structure accessible to both detector types. Together, the language results mirror the behavior observed in \cref{tab:vision_results_rescoped}: detector choice matters more when the representation is weaker or less structured, but becomes less important as modern pretrained representations scale. With strong frozen language backbones, both label-free detectors achieve high-performing OOD detection that rivals or outperforms methods using fine-tuning and class labels.

\section{Downstream Applications and Implications}

\subsection{LLM Gatekeeping and Prefill Efficiency}
\label{sec:gatekeeping}

\begin{figure}[] 
    \centering
    \begin{subfigure}[b]{0.49\textwidth}
        \centering
        \includegraphics[width=\linewidth, height=0.18\textheight, keepaspectratio]{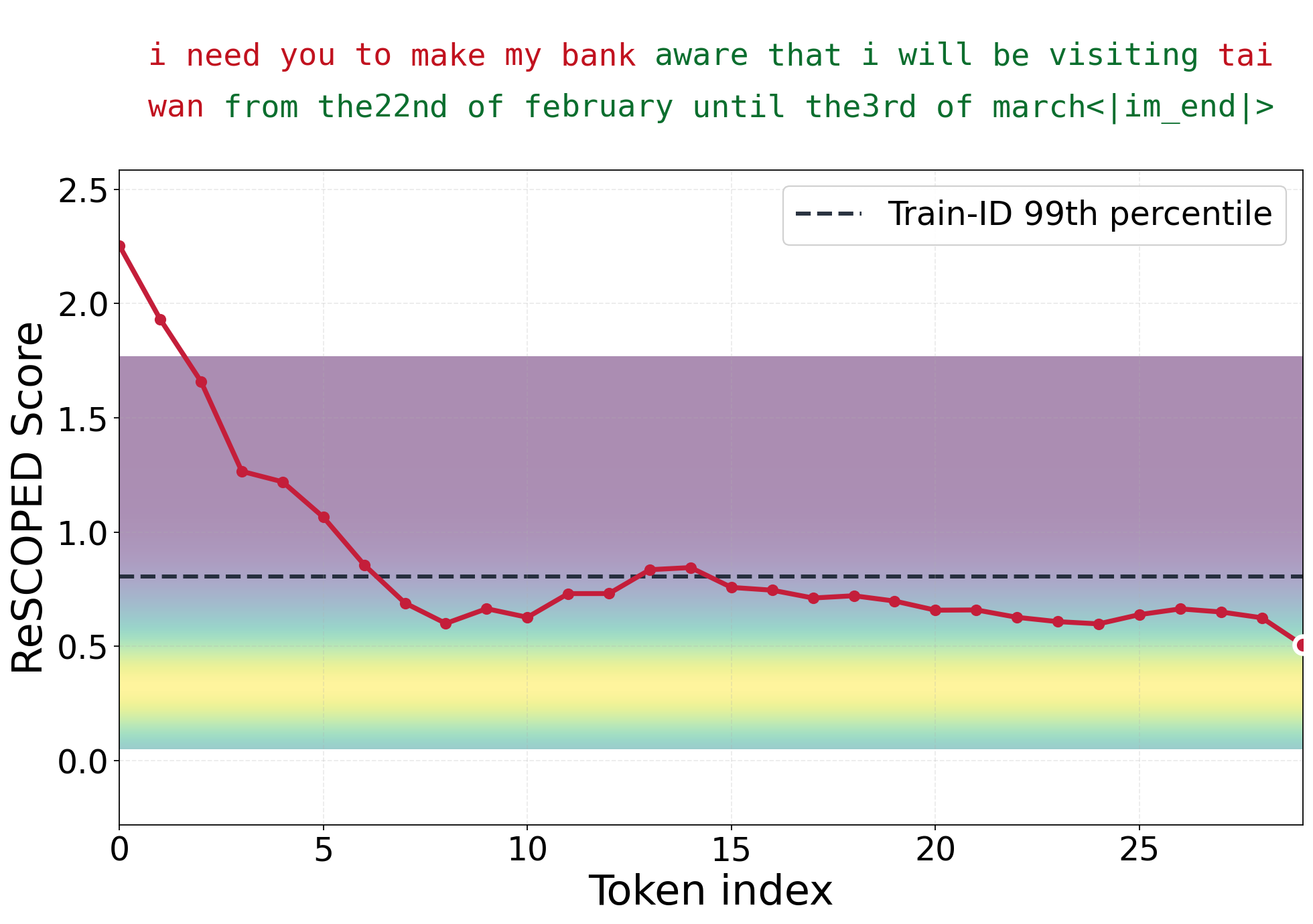}
        \caption{In-distribution utterance}
        \label{fig:id_utterance}
    \end{subfigure}
    \begin{subfigure}[b]{0.49\textwidth}
        \centering
        \includegraphics[width=\linewidth, height=0.18\textheight, keepaspectratio]{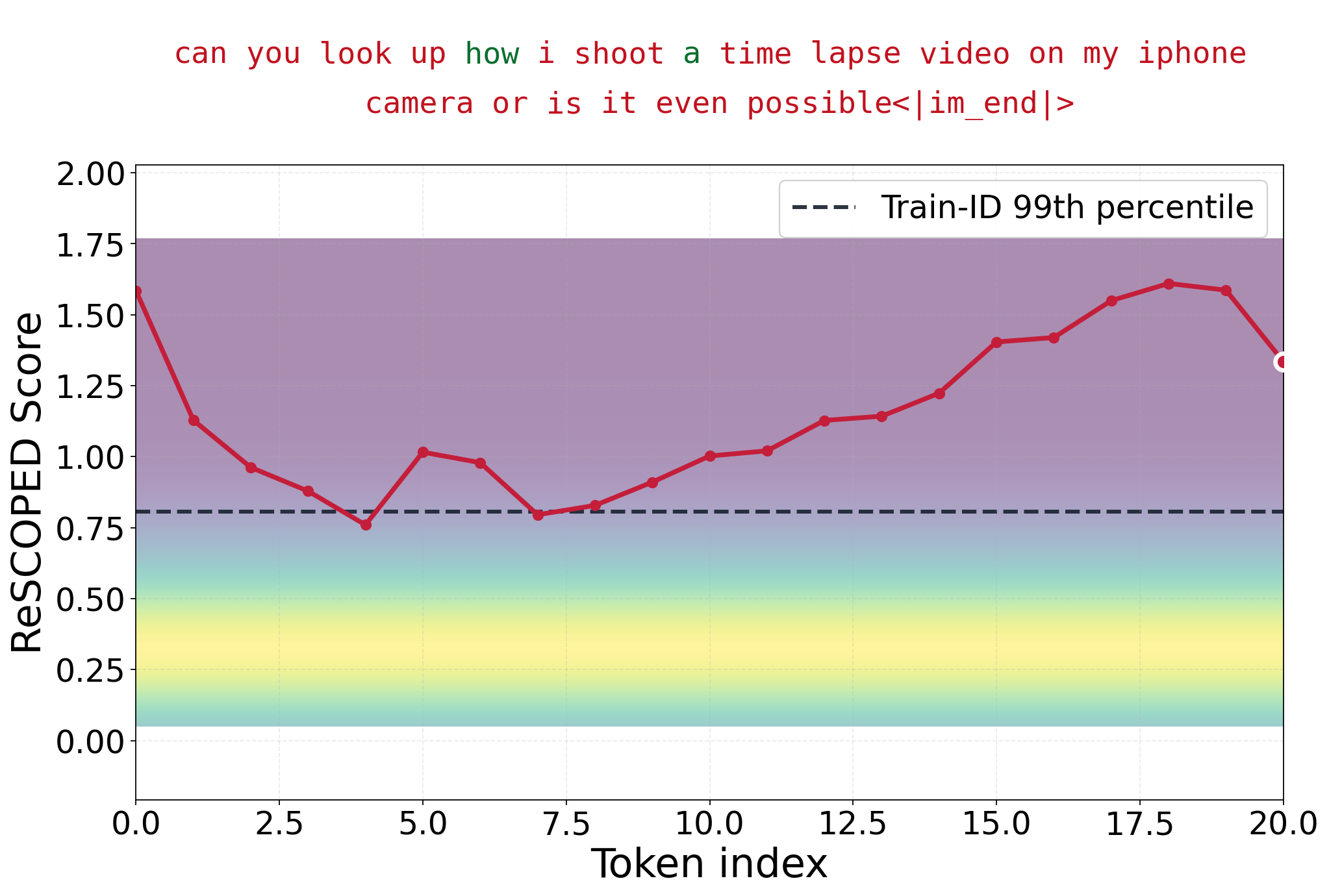}
        \caption{Out-of-distribution utterance}
        \label{fig:ood_utterance}
    \end{subfigure}
\caption{ReSCOPED prefix-level OOD scores from a frozen Qwen3-4B model on CLINC150. At each token, ReSCOPED scores the mean hidden representation of the prefix ending at that token. Token color denotes anomaly score (green lower, red higher), and the dotted line marks the 99th percentile of the in-distribution ReSCOPED scores. The in-distribution utterance falls below the threshold as context accumulates, while the OOD utterance remains above it across most of the query.}
\label{fig:gate}
\end{figure}

Representation-level OOD detection fits naturally into modern LLM inference pipelines as a gatekeeper. This deployment view is related to selective generation and abstention, where distribution-shift or reliability signals are used to decide whether a language model should proceed with generation or not \citep{ren2022out, ren2023self, azaria2023internal}. We show that an OOD detector (operating on latents, not directly on tokens) can be used during the prefill stage to decide whether to reject generation and/or route to a fallback system. Since the key-value cache is constructed from intermediate representations of the prompt, these latents can be probed without additional transformer forward passes. In \cref{fig:gate}, we test this idea with ReSCOPED, but the same prefill-stage gate can use other representation-level methods; for example, our Qwen3 results show that global Mahalanobis performs comparably in modern language representations.

At inference time, ReSCOPED adds only a single forward pass through the lightweight $\sim$8M parameter diffusion model and one Jacobian-vector product, taking approximately 5.4 ms to complete for the Qwen3-4.0B penultimate representation (2560 dimensions) on a single NVIDIA 4090 GPU. With JAX-based compilation \citep{jax2018github}, this enables immediate rejection decisions prior to autoregressive decoding, providing a lightweight gating signal at the prefill stage. In principle, multiple OOD methods could also be applied with minimal additional cost.

\cref{fig:gate} illustrates this behavior on CLINC150 using a frozen Qwen3-4B backbone. ReSCOPED is trained on full-utterance SMP representations, but at inference time we evaluate prefix-level SMP representations obtained by averaging hidden states from the beginning of the prompt through each token; the full procedure is given in Appendix \ref{app:smp_and_prefixsmp}. Early prefix scores can be unstable because they are computed from only a few tokens, but become more reliable as the prefix representation approaches the full-utterance representation used during training. Thus, ReSCOPED can provide repeated prefill-stage estimates of whether an accumulating query appears in- or out-of-distribution, suggesting a path toward decoding-stage OOD monitoring during autoregressive generation.

\subsection{Empirical Regularities Across Transformer Backbones}
\label{sec:transformergeometry}

\begin{wrapfigure}[18]{r}{0.35\textwidth}
    \vspace{-12pt}
    \centering
    \includegraphics[width=\linewidth]{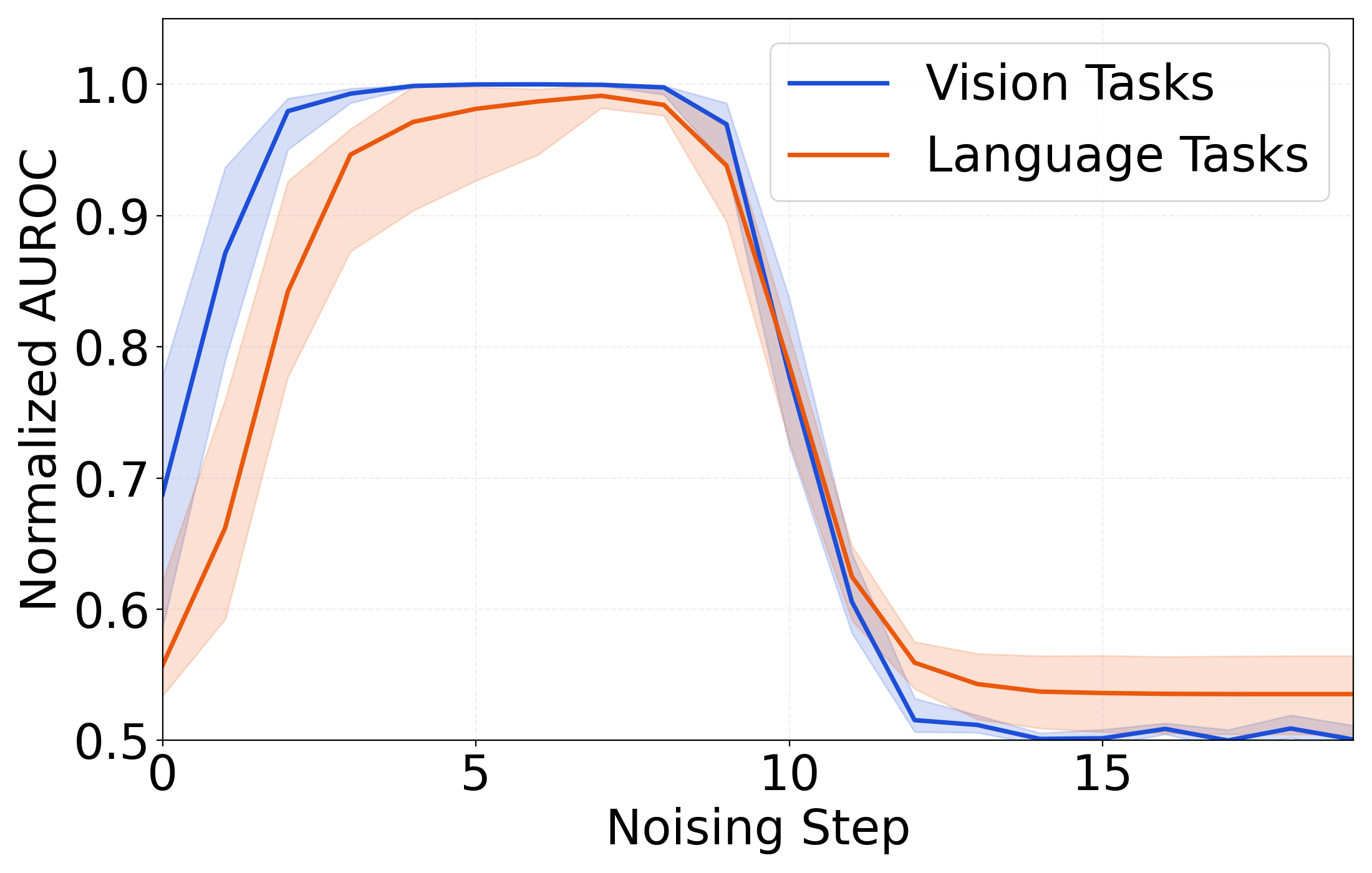}
    \caption{\textbf{OOD timestep convergence across modalities.} Interquartile mean (IQM) normalized AUROC and 95\% bootstrapped confidence intervals across 51 vision backbone-task pairings and 8 language backbone-task pairings. The aggregate curves show a broad region of highest performance.}
    \label{fig:universal}
    \vspace{-10pt}
\end{wrapfigure}

Beyond comparing detector performance, ReSCOPED probes the scale at which OOD-relevant structure is exposed in frozen representations. As shown in \cref{fig:universal}, sweeping the diffusion noise level across 59 configurations reveals a consistent performance plateau across both vision and language, with a broad highest performing region centered near $t=5$. Despite differing modalities and training objectives, OOD-relevant structure becomes accessible over a common intermediate range.

This plateau suggests that OOD separation is not tied to a task-specific noise level, but to a common geometric organization of pretrained representations. At low noise levels, ReSCOPED remains sensitive to fine-scale variation, while at high noise levels, the diffusion process removes too much structure. The intermediate regime suppresses unimportant representation variation while preserving structure useful for OOD detection.

These results provide empirical evidence for a shared granularity regime in which OOD-relevant structure in pretrained Transformer representations becomes accessible. While we do not claim universality beyond the models and datasets studied, the consistency of the plateau suggests that modern representations expose OOD-relevant information at similar levels of abstraction. Understanding the theoretical basis for this regularity remains an important direction for future work.

\section{Conclusion}
We study label-free OOD detection in frozen pretrained representations across vision and language, comparing a simple global Mahalanobis method with ReSCOPED, a local score-curvature typicality probe. ReSCOPED demonstrates that diffusion-based typicality can be moved from input space to latent space, enabling efficient OOD detection without labels, fine-tuning, or task-specific anomaly data. The surprising result is that this local expressivity does not produce a universally better detector: in weaker representations, local and global methods can outperform each other across datasets, while scaling DINOv3 and Qwen3 backbones makes the performance gap between Mahalanobis and ReSCOPED negligible across a range of tasks and modalities. Together with the synchronized denoising scale-OOD detection performance plateau and our lightweight LLM gatekeeping example, these results suggest that modern pretrained representations increasingly expose OOD-relevant geometry in a form accessible to both local and global methods. Thus, effective OOD detection in scaled foundation models may depend less on detector selection, and more on understanding when frozen representations already expose OOD-relevant geometry to methods of different construction.

\paragraph{Limitations.}
While our experiments cover a broad set of vision and language backbones, datasets, and OOD settings, they do not exhaustively cover the space of foundation models, modalities, or deployment regimes. Our comparison also focuses on two complementary label-free methods, global Mahalanobis distance and local score-curvature typicality, so future work should test whether the same representation-driven convergence holds for other detector families and harder shifts where current representations do not yet saturate performance, beyond those considered in this work.

\section*{Acknowledgments}
This work was supported by Google under a JAX AI Stack Award and by the National Science Foundation under Grant No. 2409535.

\newpage
\bibliographystyle{abbrvnat}
\bibliography{neurips_2026}
\newpage

\appendix

\section{ReSCOPED Schematic}
\label{app:rescoped_schem}
\begin{figure}[H]
    \centering
    \includegraphics[width=1.0\linewidth]{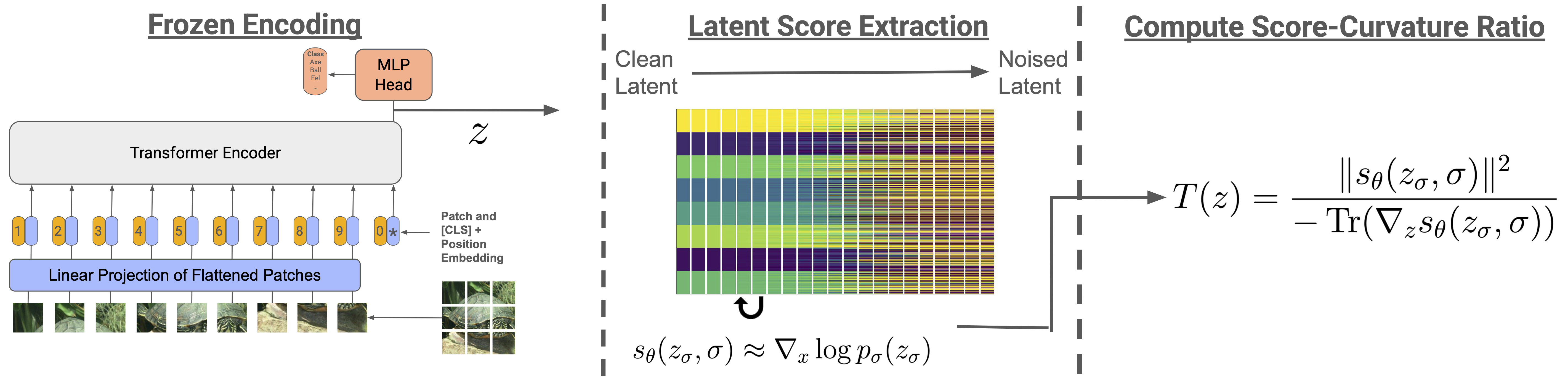}
    \caption{The ReSCOPED framework. A frozen transformer maps inputs to a latent representation $z$, where a diffusion-based score function $s_\theta$ estimates the local log-density gradient. The score-curvature ratio \citep{barkley2025scopedscorecurvatureoutofdistributionproximity} $T(z)$ is then computed to determine the typicality of the sample and determine whether it is in- or out-of-distribution.}
    \label{fig:rescoped_viz}
\end{figure}

\section{Sequence Mean Pooling (SMP) and Prefix-level SMP}
\label{app:smp_and_prefixsmp}

\paragraph{Sequence mean pooling.}
Let a tokenized input sequence have hidden states
$H = [h_1,\ldots,h_L] \in \mathbb{R}^{L \times d}$ from a frozen language backbone, with attention mask
$m \in \{0,1\}^L$ indicating non-padding tokens. Sequence mean pooling computes a single utterance representation
\[
    z_{\mathrm{SMP}}(x)
    =
    \frac{\sum_{i=1}^{L} m_i h_i}{\sum_{i=1}^{L} m_i}
    \in \mathbb{R}^{d}.
\]
This representation is then used as the input to both label-free Mahalanobis and ReSCOPED.

\paragraph{Prefix-level sequence mean pooling.}
For prefix-level scoring, let $H_{1:k}=[h_1,\ldots,h_k]$ denote the hidden states available up to prefix length $k$. We compute
\[
    z_{\mathrm{SMP}}^{(k)}(x)
    =
    \frac{\sum_{i=1}^{k} m_i h_i}{\sum_{i=1}^{k} m_i},
    \qquad k=1,\ldots,L.
\]
The resulting prefix representations can be scored by the same OOD detector used for full-sequence representations.
\section{Vision AUROC Comparisons For All Backbones, Sizes, and Datasets from \citet{heng2024out}}
\label{app:full_vision_compare}

\begin{table}[H]
\begin{center}
\caption{Performance comparison across vision in- vs out-of-distribution detection tasks as measured by AUROC score. Higher is better. \textbf{Bold} is best and \underline{underline} is second best. This set of baselines and the reporting format is reproduced from \citet{barkley2025scopedscorecurvatureoutofdistributionproximity}. As in prior work, we report computational cost as \#\texttt{F} + \#\texttt{J}, where \texttt{F} denotes a forward pass and \texttt{J} a Jacobian–vector product.}
\label{tab:vision_results_rescoped_priorwork}
\resizebox{1.0\textwidth}{!}{%
\begin{tabular}{lccccccccccc}
\toprule
\textbf{Method} & \multicolumn{3}{c}{\textbf{C10 vs}} & \multicolumn{3}{c}{\textbf{SVHN vs}} & \multicolumn{3}{c}{\textbf{CelebA vs}} & \textbf{Avg} & \textbf{\#\texttt{F} + \#\texttt{J}} \\
 & SVHN & CelebA & C100 & C10 & CelebA & C100 & C10 & SVHN & C100 & & \\
\cmidrule(lr){2-4} \cmidrule(lr){5-7} \cmidrule(lr){8-10}
Improved CD \citep{du2021improved}& 0.910 & -- & 0.830 & -- & -- & -- & -- & -- & -- & -- & -- \\
DoSE \citep{morningstar2021density} & 0.955 & 0.995 & 0.571 & 0.962 & \textbf{1.000} & 0.965 & 0.949 & 0.997 & 0.956 & 0.928 & -- \\
WAIC \citep{choi2018waic} & 0.143 & 0.928 & 0.532 & 0.802 & 0.991 & 0.831 & 0.507 & 0.139 & 0.535 & 0.601 & -- \\
TT \citep{nalisnick2019detecting} & 0.870 & 0.848 & 0.548 & 0.970 & \textbf{1.000} & 0.965 & 0.634 & 0.982 & 0.671 & 0.832 & -- \\
LR \citep{ren2019likelihood}& 0.064 & 0.914 & 0.520 & 0.819 & 0.912 & 0.779 & 0.323 & 0.028 & 0.357 & 0.524 & -- \\
\midrule
\multicolumn{12}{c}{\textbf{Diffusion-based}} \\
\midrule
NLL & 0.091 & 0.574 & 0.521 & 0.990 & \underline{0.999} & 0.992 & 0.814 & 0.105 & 0.786 & 0.652 & 1000\texttt{F} + 0\texttt{J} \\
MSMA \citep{mahmood2020multiscale} & 0.957 & \textbf{1.000} & 0.615 & 0.976 & 0.995 & 0.980 & 0.910 & 0.996 & 0.927 & 0.928 & 10\texttt{F} + 0\texttt{J} \\
DDPM-OOD \citep{graham2023denoising} & 0.390 & 0.659 & 0.536 & 0.951 & 0.986 & 0.945 & 0.795 & 0.636 & 0.778 & 0.742 & 350\texttt{F} + 0\texttt{J} \\
LMD \citep{liu2023unsupervised} & 0.992 & 0.557 & 0.604 & 0.919 & 0.890 & 0.881 & 0.989 & \textbf{1.000} & 0.979 & 0.868 & $10^4$\texttt{F} + 0\texttt{J} \\
\midrule
\multicolumn{12}{c}{\textbf{Curvature and Diffusion-Based}} \\
\midrule
DiffPath \citep{heng2024out} & 0.910 & 0.897 & 0.590 & 0.939 & 0.979 & 0.953 & 0.998 & \textbf{1.000} & \textbf{0.998} & 0.918 & 10\texttt{F} + 0\texttt{J} \\
SCOPED \citep{barkley2025scopedscorecurvatureoutofdistributionproximity} & 0.814 & 0.940 & 0.477 & 0.971 & 0.996 & 0.959 & 0.925 & 0.994 & 0.962 & 0.892 & 2\texttt{F} + 2\texttt{J} \\
\midrule
\midrule

\textbf{Unsupervised Mahalanobis DINOv1 ViT-B} & 0.761 & 0.998 & 0.895 & 1.000 & 1.000 & 1.000 & 1.000 & 1.000 & 0.997 & 0.961 & \textbf{1\texttt{F} + 0\texttt{J}} \\
\textbf{Unsupervised Mahalanobis DINOv2 ViT-B} & 0.829 & 1.000 & 0.952 & 1.000 & 1.000 & 1.000 & 1.000 & 0.999 & 0.997 & 0.975 & \textbf{1\texttt{F} + 0\texttt{J}} \\
\textbf{Unsupervised Mahalanobis DINOv3 ViT-B} & 0.990 & 1.000 & 0.923 & 1.000 & 1.000 & 1.000 & 1.000 & 1.000 & 0.997 & 0.990 & \textbf{1\texttt{F} + 0\texttt{J}} \\
\textbf{ReSCOPED DINOv1 ViT-B} & 0.683 & 0.985 & 0.819 & 1.000 & 1.000 & 1.000 & 1.000 & 0.999 & 0.996 & 0.942 & \underline{2\texttt{F} + 1\texttt{J}} \\
\textbf{ReSCOPED DINOv2 ViT-B} & 0.698 & 0.991 & 0.894 & 1.000 & 1.000 & 0.999 & 1.000 & 0.995 & 0.996 & 0.953 & \underline{2\texttt{F} + 1\texttt{J}} \\
\textbf{ReSCOPED DINOv3 ViT-B} & 0.968 & 0.996 & 0.919 & 1.000 & 1.000 & 1.000 & 1.000 & 1.000 & 0.996 & 0.986 & \underline{2\texttt{F} + 1\texttt{J}} \\

\textbf{Unsupervised Mahalanobis ViT-B} & 0.942 & 0.998 & 0.869 & 0.999 & 1.000 & 0.998 & 1.000 & 0.998 & 0.996 & 0.989 & \textbf{1\texttt{F} + 0\texttt{J}} \\
\textbf{Label-Free Mahalanobis Dinov3 ViT-S} & \textbf{0.997} & 1.000 & 0.864 & 1.000 & 1.000 & 1.000 & 1.000 & 1.000 & 0.998 & 0.984 & \textbf{1\texttt{F} + 0\texttt{J}} \\
\textbf{Label-Free Mahalanobis Dinov3 ViT-B} & 0.990 & 1.000 & 0.923 & 1.000 & 1.000 & 1.000 & 1.000 & 1.000 & 0.997 & 0.990 & \textbf{1\texttt{F} + 0\texttt{J}} \\
\textbf{Label-Free Mahalanobis Dinov3 ViT-L} & 0.983 & 1.000 & \textbf{0.970} & 1.000 & 1.000 & 1.000 & 1.000 & 0.999 & 0.997 & \textbf{0.994} & \textbf{1\texttt{F} + 0\texttt{J}} \\

\textbf{ReSCOPED ViT-B} & 0.885 & 0.987 & 0.759 & \underline{0.998} & \underline{0.999} & \underline{0.997} & \underline{0.999} & \underline{0.999} & 0.995 & 0.958 & \underline{2F + 1J} \\
\textbf{ReSCOPED DINOv3 ViT-S} & \underline{0.994} & \underline{0.999} & 0.864 & \textbf{1.000} & \textbf{1.000} & \textbf{1.000} & \underline{0.999} & \textbf{1.000} & 0.993 & 0.983 & \underline{2F + 1J} \\
\textbf{ReSCOPED DINOv3 ViT-B} & 0.968 & 0.996 & 0.919 & 1.000 & 1.000 & 1.000 & 1.000 & 1.000 & 0.996 & 0.986 & \underline{2\texttt{F} + 1\texttt{J}} \\
\textbf{ReSCOPED DINOv3 ViT-L} & 0.969 & 0.997 & \underline{0.969} & \textbf{1.000} & \textbf{1.000} & \textbf{1.000} & \textbf{1.000} & 0.997 & 0.995 & \underline{0.992} & \underline{2F + 1J} \\
\bottomrule
\end{tabular}%
}
\end{center}
\end{table}
\section{Label-free Mahalanobis and ReSCOPED Hyperparameters}
\label{app:hparams}

\subsection{Global Label-free Mahalanobis}

\begin{table}[h]
\centering
\caption{Hyperparameters for global label-free Mahalanobis.}
\label{tab:maha_hparams}
\resizebox{0.55\textwidth}{!}{%
\begin{tabular}{ll}
\toprule
\textbf{Quantity} & \textbf{Value} \\
\midrule
Input data & Unlabeled ID training latents \\
Covariance ridge $\lambda$ & $10^{-4}$ \\
Evaluation & ID test vs. OOD test \\
\bottomrule
\end{tabular}
}
\end{table}

Given ID training latents $\{z_i\}_{i=1}^{n}$, we estimate
\[
    \mu = \frac{1}{n}\sum_{i=1}^{n} z_i,
    \qquad
    \Sigma_{\mathrm{global}}
    =
    \frac{1}{n}
    \sum_{i=1}^{n}
    (z_i-\mu)(z_i-\mu)^\top
    + \lambda I,
\]
with $\lambda=10^{-4}$. The global Mahalanobis score is
\[
    S_{\mathrm{global}}(z)
    =
    (z-\mu)^\top \Sigma_{\mathrm{global}}^{-1}(z-\mu).
\]
Larger values indicate greater distance from the fitted ID representation distribution. 

\subsection{ReSCOPED}

\begin{table}[h]
\centering
\caption{Hyperparameters for ReSCOPED.}
\label{tab:rescoped_hparams}
\resizebox{0.55\textwidth}{!}{%
\begin{tabular}{ll}
\toprule
\textbf{Quantity} & \textbf{Value} \\
\midrule
Input data & Unlabeled ID training latents \\
Training steps & $150{,}000$ \\
Batch size & $256$ \\
Optimizer & Adam \\
Learning rate & $3\times 10^{-4}$ with cosine decay \\
Denoiser & Residual MLP, width $1024$, $6$ layers \\
Time embedding dimension & $256$ \\
Scoring noise level & $\sigma \approx 0.099$ \\
KDE bandwidth & $0.2$ \\
Evaluation & ID test vs. OOD test \\
\bottomrule
\end{tabular}
}
\end{table}

ReSCOPED trains a latent diffusion observer on ID training latents. The EDM normalizer is fit on ID training latents and reused for scoring. At evaluation time, we use a fixed noise level $\sigma \approx 0.099$.

Let $s_\theta(z,\sigma)$ denote the learned diffusion score. ReSCOPED computes
\[
    T(z;\sigma)
    =
    \frac{\|s_\theta(z,\sigma)\|_2^2}
    {-\operatorname{Tr}\!\left(\nabla_z s_\theta(z,\sigma)\right) + \epsilon},
    \qquad \epsilon=10^{-8}.
\]
We fit a one-dimensional Gaussian-kernel KDE to ID training values $\{T_i\}_{i=1}^{n}$ with bandwidth $h=0.2$. 

\section{Frozen Pretrained Models Used in This Work}
\label{app:models}

\begin{table}[h]
\centering
\caption{Frozen pretrained backbones and latent representations used by the OOD detectors.}
\label{tab:frozen_models}
\resizebox{0.95\textwidth}{!}{%
\begin{tabular}{llll}
\toprule
\textbf{Backbone} & \textbf{Checkpoint} & \textbf{Inference} & \textbf{Latent representation} \\
\midrule
DINOv3 ViT-S/B/L 
& \texttt{facebook/dinov3-\{vits16,vitb16,vitl16\}-pretrain-lvd1689m}
& Bonsai/JAX
& post-norm CLS/global image vector \\
ViT-B/16 
& \texttt{google/vit-base-patch16-224}
& Bonsai/JAX
& post-norm CLS vector before classifier \\
DINOv2 ViT-B 
& \texttt{facebook/dinov2-base}
& Hugging Face Transformers
& CLS/global image vector \\
DINOv1 ViT-B/16 
& \texttt{facebook/dino-vitb16}
& Hugging Face Transformers
& CLS/global image vector \\
BERT 
& \texttt{bert-base-uncased}
& Hugging Face Transformers
& mean-pooled token hidden states \\
Qwen3 
& \texttt{Qwen/Qwen3-\{0.6B,1.7B,4B\}}
& Bonsai/JAX
& mean-pooled non-padding token hidden states \\
\bottomrule
\end{tabular}
}
\end{table}

\paragraph{Latent extraction.}
For each modality, we extract a single frozen embedding $\phi(x)$ and use it as the input to both label-free Mahalanobis and ReSCOPED. The DINOv3, ViT-B/16, and Qwen3 checkpoints are sourced from Hugging Face but evaluated through Bonsai/JAX \citep{bonsai2024google}. DINOv1, DINOv2, and BERT are evaluated through Hugging Face \texttt{transformers} \citep{wolf2019huggingface}. For vision models, we use a CLS-style global image representation. In the Bonsai/JAX ViT and DINOv3 models, this corresponds to the post-final-layer-normalization CLS token; for DINOv3, this vector is exposed through the pooled-output field. For BERT, we compute an utterance embedding by sequence mean-pooling token hidden states. For Qwen3, we compute an utterance embedding by sequence mean-pooling non-padding token hidden states. Thus, all backbones provide one fixed-length representation per input, but the pooling mechanism differs across modalities: vision uses a designated global token, while language uses explicit sequence mean pooling over token states.

\section{Datasets and Preprocessing}
\label{app:datasets}

\paragraph{Small-scale vision datasets.}
We use \texttt{torchvision.datasets} with canonical train/test splits for CIFAR-10, CIFAR-100, SVHN, and CelebA. Mahalanobis and ReSCOPED are fit using ID train latents and evaluated on ID test versus OOD test.

\paragraph{Vision preprocessing.}
Images are converted to RGB, resized to the model input resolution, scaled to $[0,1]$, and normalized with ImageNet mean and standard deviation. For CelebA in the small-scale vision comparisons, images are first resized to $32\times 32$ before being resized to the model input resolution, matching the comparison protocol of \citet{heng2024out}.

\paragraph{OpenOOD ImageNet-200.}
For ImageNet-200 experiments, we use the OpenOOD image lists for ImageNet-200 ID train/test and SSB-hard or NINCO OOD test sets. Images are decoded to RGB, resized to $224\times224$, and ImageNet-normalized. Detectors are fit on ImageNet-200 train latents and evaluated on ImageNet-200 test versus OOD test.

\paragraph{CLINC150.}
We use \texttt{datasets.load\_dataset("clinc\_oos", "plus")}. In-domain train examples are used for fitting the detector, and evaluation is performed on in-domain test examples versus OOS test examples. The validation split is not used for detector training or reported test metrics.

\paragraph{SST cross-corpus.}
For SST cross-corpus evaluation, SST train examples are used for fitting, SST test examples are used as ID test data, and OOD examples are drawn from the specified cross-corpus sources. We sample $500$ examples from each OOD source with random seed $42$ as in \citet{xu2021unsupervised}.

\section{Per Backbone Reproduction of \cref{fig:universal} Without Normalization}
\begin{figure}[H]
    \centering
    \includegraphics[width=0.5\linewidth]{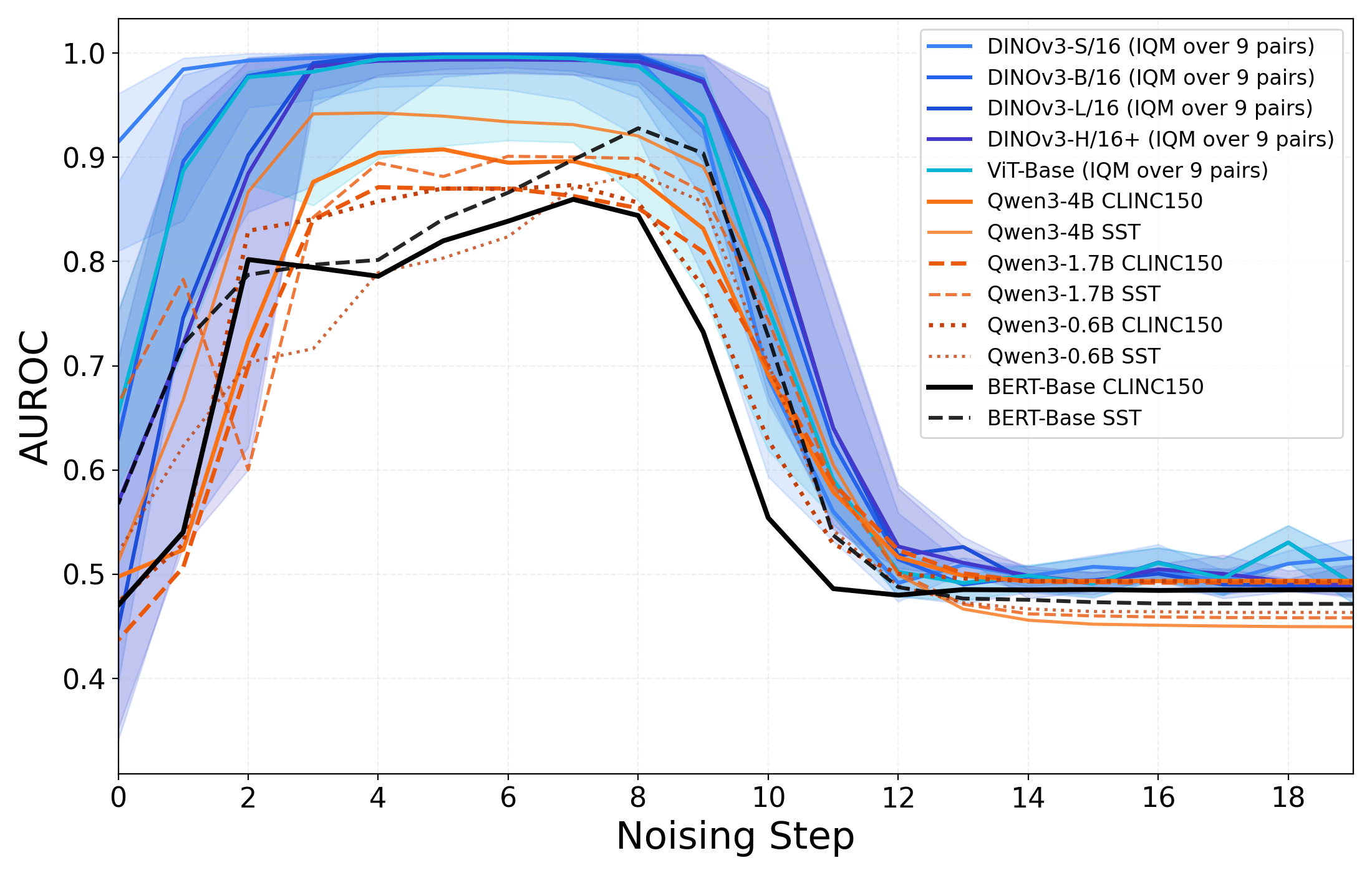}
    \caption{AUROC scores for all of the backbones that comprise \cref{fig:universal}.}
    \label{fig:placeholder}
\end{figure}


\newpage

\end{document}